\documentclass{article}

\usepackage[final]{neurips_2023_ml4ad}

\usepackage[utf8]{inputenc} 
\usepackage[T1]{fontenc}    
\usepackage{inconsolata}

\usepackage{algpseudocode}  
\usepackage[ruled, linesnumbered, boxed]{algorithm2e}
\usepackage{parskip}
\usepackage{changepage}

\usepackage{color}
\usepackage{url}            
\usepackage{wrapfig}
\usepackage{caption}
\usepackage{cancel}
\usepackage{graphicx}
\usepackage{subfig}
\usepackage{epsfig} 
\usepackage{times} 
\usepackage{booktabs}
\usepackage{booktabs}       
\usepackage{amsmath}
\usepackage{amsthm}
\usepackage{amssymb}  
\usepackage[table, dvipsnames, svgnames, x11names]{xcolor} 
\usepackage{marvosym}
\definecolor{mygray}{gray}{.9}
\usepackage{pifont}
\usepackage{nicefrac}       
\usepackage{microtype}      
\usepackage[table, dvipsnames, svgnames, x11names]{xcolor}        
\usepackage{colortbl}

\usepackage{multirow}
\usepackage{multicol}
\usepackage{cite}

\usepackage{enumitem}
\usepackage{graphics} 
\usepackage{svg}
\usepackage{bm}
\usepackage{float} 
\usepackage{stfloats}
\usepackage{adjustbox}
\usepackage{tablefootnote}
\usepackage[colorlinks,
linkcolor=blue,
anchorcolor=blue,
citecolor=blue,
urlcolor=Aqua,
backref=page]{hyperref}
\usepackage{wrapfig}

\title{Stackelberg Driver Model for Continual Policy Improvement in Scenario-Based Closed-Loop Autonomous Driving}

\author{
Haoyi Niu$^{1}$\thanks{Work done with equal contribution.}\hspace{1.5mm}, Qimao Chen$^{1*}$, Yingyue Li$^{1}$, Yi Zhang$^{1}$, Jianming Hu$^{1, 2}$\thanks{Corresponding author.}\\
$^1$ Tsinghua University, Beijing, China\\
$^2$ Beijing National Research Center for Information Science and Technology, Beijing, China\\
\texttt{\{nhy22, cqm20\}@mails.tsinghua.edu.cn}\\
\texttt{hujm@mail.tsinghua.edu.cn}
\thanks{Work supported by National Natural Science Foundation of China under Grant No.62333015.}
}

\begin{document}
	
\maketitle
\begin{abstract}
The deployment of autonomous vehicles (AVs) has faced hurdles due to the dominance of rare but critical corner cases within the long-tail distribution of driving scenarios, which negatively affects their overall performance. To address this challenge, adversarial generation methods have emerged as a class of efficient approaches to synthesize safety-critical scenarios for AV testing. However, these generated scenarios are often underutilized for AV training, resulting in the potential for continual AV policy improvement remaining untapped, along with a deficiency in the closed-loop design needed to achieve it. Therefore, we tailor the Stackelberg Driver Model (SDM) to accurately characterize the hierarchical nature of vehicle interaction dynamics, facilitating iterative improvement by engaging background vehicles (BVs) and AV in a sequential game-like interaction paradigm. With AV acting as the leader and BVs as followers, this leader-follower modeling ensures that AV would consistently refine its policy, always taking into account the additional information that BVs play the best response to challenge AV. Extensive experiments have shown that our algorithm exhibits superior performance compared to several baselines especially in higher dimensional scenarios, leading to substantial advancements in AV capabilities while continually generating progressively challenging scenarios.
\end{abstract}
	
\section{Introduction}\label{Introduction} 

Autonomous vehicles (AVs) have sparked a global wave of hope and potential as their capability continues to advance and evolve. However, further progress toward the ultimate goal of real-world deployment is consistently impeded by the intractable long-tail distribution of naturalistic driving scenarios, within which a handful of homogeneous and unchallenging scenarios comprise the majority, while critical and risky ones are severely limited and underrepresented~\citep{DRL, cui2019class, ding2020learning, kou2008worst, o2018scalable}. 
To address this data imbalance issue, several attempts have been made to generate rare and risky scenarios~\citep{Generating_adversarial_Scenarios, Advsim, King, re2h2o,ding2023survey}, which pose significant challenges to AVs. 
Subsequently, these scenarios can compensate for the lack of rare-seen scenarios in naturalistic driving data (NDD) and are extensively employed to accelerate the industrial AV testing phase~\citep{zhao2016accelerated, zhao2017accelerated,chen2018model,ding2020learning,feng2021intelligent,DRL}. 
In this process, background vehicles (BVs) are continuously updated to surface situations where AV encounters failures. The tested AV, however, remains a fixed agent throughout this iterative procedure.
This leads to scenarios not being fully exploited and their value not being fully harnessed.
Correspondingly, they fail to establish a closed-loop system that seamlessly integrates scenario-based AV training and testing for continual driving policy optimization.



\begin{figure}[t]
    \centering
    \includegraphics[width=1\linewidth]{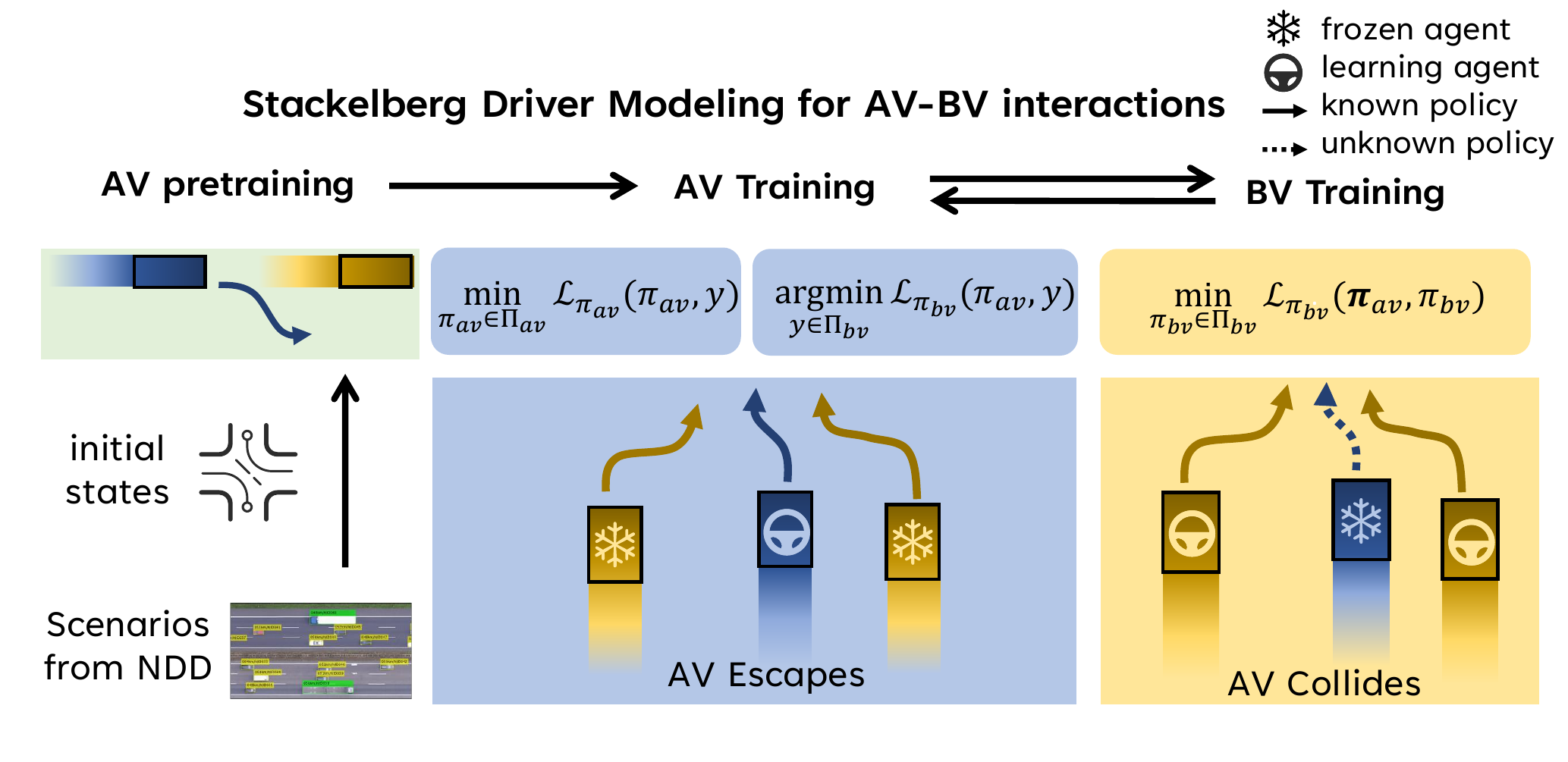}
    \vspace{-15pt}
    \caption{\textbf{Stackelberg Driver Model}. We first pretrain AV agent and then employ the Stackelberg game to characterize AV-BV interaction, where AV acts as the leader and BVs act as the followers.}
    \label{fig:arch}
    \vspace{-28pt}
\end{figure}

We believe that these generated scenarios can further be leveraged to optimize AV policy~\citep{RARL, Xu2022TrustworthyRL}. 
With AV-BV interactions modeled through an adversarial game process, we can ingeniously achieve continual driving performance improvement by optimizing AV policy over dynamically generated adversarial scenarios.
However, commonly adopted simultaneous game modeling falls short in fully characterizing the \textbf{hierarchical relationships in interactions between AV and BVs}, in which BVs are just considered as auxiliary variables within the environment to serve the ultimate goal of AV improvement.
Alternatively, the Stackelberg Game~\citep{von2010market}, being one of the fundamental methods in sequential game theory, has demonstrated favorable efficiency and convergence properties in task environments that exhibit hierarchical order of play~\citep{implicit_ld_in_sg, Convergence_of_LD_in_SG}.
By prioritizing one of the players (leader) with information about the others (followers), the leader gains an advantage in the bi-level optimization formulation~\citep{BLO}. 
Specifically, Stackelberg games find extensive applications in situations that involve distinct leader-follower relationships~\citep{joel2002actor}. 
In common practice, Stackelberg Generative Adversarial Networks (GANs)~\citep{implicit_ld_in_sg} typically designate the generator as the leader and the discriminator as the follower. As another application, Stackelberg Actor-Critic~\citep{zheng2022stackelberg} assigns the actor as the leader and the critic as the follower. 
In principle, the dominant player that aligns more with the goal of the task assumes the role of leader, taking advantage of additional decision information of the followers, with the objective of strategically lowering its own payoff while aiming to achieve superior performance overall~\citep{von2010market}.

Intuitively, we observe that a similar hierarchical leader-follower relationship exists between AV and BVs in the context of autonomous driving. 
To capture this feature, we naturally propose the novel \textbf{Stackelberg Driver Model (SDM)} framework where AV assumes the role of the leader and BVs act as the followers.
In this closed-loop continual optimization framework, AV policy is first updated with the safety-critical scenarios produced by BV policy, knowing that BV would play the best response. On the other hand, BVs are then optimized to risk AV adversarially over time yet without awareness of the current AV reaction. 
The whole architecture and the optimization mechanism of SDM framework are elaborated in Fig.~\ref{fig:arch}. 
The contributions of this work are two-fold: 
(1) We stress the necessity of constructing a closed-loop continual policy improvement framework for scenario-based autonomous driving, integrating previously used testing scenarios into the training phase;
(2) We introduce the SDM as a novel approach that aligns with the hierarchical nature of interactions between AV and BVs, facilitating AV performance improvement in a leader-follower order of play. 
We empirically demonstrate that AV and BV agents can both achieve better performance during the game with SDM. Essentially, SDM outperforms those competing baselines with simple adversarial training paradigms, non-game modeling and other game modeling approaches, especially in higher dimensional scenarios.



\section{Related Work}\label{rw}
\subsection{Adversarial Learning in Autonomous Driving}\label{rw-1}

Using adversarial learning in autonomous driving is a promising way to generate more challenging scenarios for testing, where AV belongs to the environment and the generator (BV) is the agent we can control~\citep{ding2023survey}. Feng et al. ~\citep{feng2021intelligent, corner} use Deep Q-Network to generate discrete adversarial traffic scenarios. Wachi ~\citep{wachi2019failure} uses Multi-agent DDPG~\citep{multiDDPG} to control two surrounding vehicles to attack the ego vehicle. DR2L~\citep{niu2021dr2l} uses an adversarial model to robustify autonomous vehicles by surfacing harder events via domain randomization. 
$\text{(Re)}^2\text{H2O}$~\citep{re2h2o} presents Hybrid Offline-and-Online Reinforcement Learning to generate safety-critical scenarios. STRIVE~\citep{STRIVE} proposes a framework that generates safety-critical scenarios via gradient-based optimization.
Adversarial learning approach ensures that the generated scenarios are varied, adversarial, and critical for validating AV safety~\citep{ding2023survey}. However, these generated scenarios are not fully utilized to train AV in a continual learning manner in these previous works. Research has been conducted on adversarial RL environments that are evolving~\citep{wang2020enhanced, wang2019poet}, but only evaluated in simplified environments such as bipedal walkers or single adversarial agent, which does not meet the needs of autonomous driving. Therefore, in our work, we employ adversarial learning to generate challenging scenarios containing multiple BVs for \textbf{training AV} so that we can establish a closed-loop framework for autonomous driving.

\subsection{Stackelberg Game}\label{rw-2}
Stackelberg game is a sequential structure of play, of which the simplest form involves a leader (the first player) and a follower (the second player) ~\citep{stackelberggame}. The leader first commits to a strategy on the premise of knowing the strategy function of the follower, which means that the leader knows the best response of the follower~\citep{huang2022robust}. Correspondingly, the follower chooses a strategy to maximize its own payoff while adapting to the leader's policy. This process is repeated until a \textit{Stackelberg Equilibrium}~\citep{implicit_ld_in_sg} (a specific Nash Equilibrium~\citep{kreps1989nash, holt2004nash}) is reached~\citep{spg, Convergence_of_LD_in_SG}. 
Learning dynamics in Stackelberg games ensure that leaders can fully utilize this leader-follower structure to benefit their own utility and potentially improve overall efficiency~\citep{liu1998stackelberg, SeqAndSim}. 

Due to such characteristics, Stackelberg game is regularly applied to find the Nash Equilibria of bi-level optimization objectives through its stable convergence~\citep{zhou2019survey}. For example, in Stackelberg Actor Critic~\citep{zheng2022stackelberg} structure, deeming actor as the leader and critic as the follower significantly outperforms the standard actor-critic~\citep{konda1999actor,joel2002actor} algorithm counterparts.
Several researches applied the idea of Stackelberg game to the architectural design of GAN~\citep{GAN}, proposing an improved GAN architecture where the generator (leader) optimizes a cost function that depends on parameters of the discriminator (follower)~\citep{implicit_ld_in_sg}.

Despite its good performance, little prior research has used Stackelberg game to model the interactions between AV and BVs in autonomous driving domain. In this paper, we explore the potential of Stackelberg game by training AV and BV sequentially and adversarially. We show evidence that the excellent convergence and stability properties of Stackelberg game enable BV to generate more critical scenarios that are specifically designed for AV. AV is then trained based on this tailored BV environment, thereby enabling faster convergence and performance improvement. 

\section{Methodology}\label{method}

\subsection{Problem Formulation}\label{3-2}
The interactions between driving agents and the environment can be conceptualized as a Markov game~\citep{littman1994markov} presented by a tuple $(S, A, R, P, \rho, \gamma)$, where $S$, $A$, $R$, $P$, $\rho$ denote the state, action, reward space, transition probability and initial state distribution, respectively. 
We assume that the traffic environment involves an AV denoted as $V_0$ and a set of $N$ BVs denoted as $V_i, i = 1, \dots, N$.
For each vehicle $i$ at moment $t$, the state vector $\mathbf{s}_t^i=[x_t^i, y_t^i, v_t^i, \theta_t^i]$, where $(x,y)$ is lateral and longitudinal position , $v$ is longitudinal velocity and $\theta$ is heading angle. The action vector $\mathbf{a}_t^i=[\Delta{v_t^i}, \Delta{\theta_t^i}]$, where $\Delta{v_t^i}$ and $\Delta{\theta_t^i}$ are the increment of longitudinal velocity  and heading angle between two consecutive moments, respectively. 
During the game, a state at time step $t$ encapsulates the states of all traffic participants at this time step:
$
\mathbf{s}_t=[\mathbf{s}_t^0,\mathbf{s}_t^1,\dots,\mathbf{s}_t^{N}]^T
$.  State transitions are controlled by the current state and one action from each agent:
$
T: S\times A_0 \times A_1 \times \cdots A_N \rightarrow P(S)
$. Each agent $i$ also has a reward function:
$
R_i: S\times A_0 \times A_1 \times \cdots A_N \rightarrow \mathbb{R} 
$. The conditional probability form of transition probability and reward functions are
$
p(\mathbf{s}_{t+1}|\mathbf{s}_t,\mathbf{a}_0, \dots, \mathbf{a}_N)
$
and
$
r_i (\mathbf{s}_{t}, \mathbf{a}_0, \dots, \mathbf{a}_N, \mathbf{s}_{t+1})
$
respectively.


The design of the AV reward function at moment $t$, $r_{t, av}$, involves concerns on AV speed ($r_{t, av}^{speed}$) and AV collision occurrence ($r_{t, av}^{col}$):
\begin{equation}
r_{t,av}=r_{t, av}^{speed}+r_{t, av}^{col}
\end{equation}
To increase the driving speed of AV, we reward AV for higher velocity: $r_{t, av}^{speed} = \frac{v_{t,av}}{v_{max,av}}$.
Naturally, AV agent deserves penalty for AV collision at the last moment in the scenario, thus avoiding AV failure cases. Therefore, $r_{t, av}^{col}=-R_a$ if AV collides with BV else $0$.

For BV, the reward function, $r_{t, bv}$, should additionally includes the collision rate between BVs ($r_{t, bv}^{col}$):
\begin{equation}
r_{t,bv}=-r_{t, av}^{speed}-r_{t, av}^{col}+r_{t, bv}^{col}
\end{equation}
where $r_{t,bv}^{col}=-R_b$ if BV collides with BV else $0$.
The aim is for BV to create more and more difficult scenarios for AV with action $\mathbf{a}_{t,bv}=[\mathbf{a}_t^1, \dots, \mathbf{a}_t^N] \in A_{bv}$, while allowing AV to learn and develop a stronger policy $\pi_{av}$ performing $\mathbf{a}_{t,av}=[\mathbf{a}_t^0] \in A_{av}$ to handle the challenges posed by BV. 
The objective of the optimization problem is to maximize the cumulative rewards $r_{t,av}=r_{av}(\mathbf{s}_t, \mathbf{a}_{t,av}, \mathbf{a}_{t,bv}) \in R_{av}$ and $r_{t,bv}=r_{bv}(\mathbf{s}_t, \mathbf{a}_{t,bv}, \mathbf{a}_{t,av}) \in R_{bv}$, which are discounted by the same $\gamma$, given state $\mathbf{s}_t \in S$:
\begin{equation}\label{eq4}
\begin{aligned}
&J_{av}\left(P_{av},\pi_{av}\right)=
\mathbb{E}_{\substack{\mathbf{s}_0 \in \rho, \mathbf{s}_{t+1}\sim P_{av}(\cdot|\mathbf{s},\mathbf{a}_{t,av},\mathbf{a}_{t,bv})\\
\mathbf{a}_{t,av} \sim \pi_{av}(\cdot|\mathbf{s}_t),
\mathbf{a}_{t,bv} \sim \pi_{bv}(\cdot|\mathbf{s}_t)}}\left[\sum_{t=0}^{H}\gamma^t r_{t,av}\right]\\
&J_{bv}\left(P_{bv},\pi_{bv}\right)=
\mathbb{E}_{\substack{\mathbf{s}_0 \in \rho, \mathbf{s}_{t+1}\sim P_{bv}(\cdot|\mathbf{s},\mathbf{a}_{t,bv},\mathbf{a}_{t,av})\\
\mathbf{a}_{t,bv} \sim \pi_{bv}(\cdot|\mathbf{s}_t),
\mathbf{a}_{t,av} \sim \pi_{av}(\cdot|\mathbf{s}_t)}}\left[\sum_{t=0}^{H}\gamma^t r_{t,bv}\right]
\end{aligned}
\end{equation}

In conventional Actor-Critic formalism ~\citep{SAC}, Q-function $\hat{Q}$ is approximated by minimizing the standard Bellman error (Eq.~\ref{eq11}), while policy $\hat{\pi}$ is optimized by maximizing the Q-function (Eq.~\ref{eq12}):
\begin{equation}\label{eq11}
\hat{Q}\leftarrow\arg\min_Q\mathbb{E}_{\mathbf{s},\mathbf{a},\mathbf{s}^{\prime}\sim\mathcal{U}}\left[\frac12\left((Q-\hat{\mathcal{B}}^{\boldsymbol{\pi}}\hat{Q})(\mathbf{s},\mathbf{a})\right)^2\right]
\end{equation}
\begin{equation}\label{eq12}
\hat{\boldsymbol{\pi}}\leftarrow\arg\max_\pi\mathbb{E}_{\mathbf{s},\mathbf{a}\sim\mathcal{U}}\left[\hat{Q}(\mathbf{s},\mathbf{a})\right]
\end{equation}
where $\mathcal{U}$ denotes the data buffer generated by a previous version of policy $\hat{\pi}$ through online simulation interactions. The Bellman evaluation operator $\hat{\mathcal{B}}^{\pi}$ is given by $\hat{\mathcal{B}}^{\pi}\hat{Q}(\mathbf{s},\mathbf{a})=r(\mathbf{s},\mathbf{a})+\gamma\mathbb{E}_{\mathbf{a}^{\prime}\sim\hat{\boldsymbol{\pi}}(\mathbf{a}^{\prime}|\mathbf{s}^{\prime})}\left[\hat{Q}\left(\mathbf{s}^{\prime},\mathbf{a}^{\prime}\right)\right]$. 
We model the stochasticity by using a Gaussian distribution $\mathcal{N}(\mu,\sigma)$ for both AV and BV policies, with $\mu$ and $\sigma$ approximated by a neural network $\mathcal{W}_{\theta}$~\citep{re2h2o}. 
By utilizing a different neural network, the Q-function can be approximated as $\hat{Q}_{\phi}$. Subsequently, through optimizing the policy network of both AV and BV with SDM (Sec.~\ref{3-3}), the objectives in Eq.~\ref{eq4} can be maximized, leading to a BV that is more aggressive and higher-quality, as well as an AV that is both secure and efficient. 
\subsection{Stackelberg Driver Model for AV-BV Interactions}\label{3-3}
For the Stackelberg AV-BV model, the leader of the game knows the other players' response and incorporates this response in its policy update. In the optimization of Eq.~\ref{eq4}, the policy of AV serves as the ultimate optimization objective. As a result, AV policy network (here we use \textbf{Actor-Critic}~\citep{konda1999actor} architecture) should possess complete knowledge of the policies of BVs. Consequently, we designate AV as the \textbf{leader} and BV as the \textbf{follower} in the decision-making hierarchy. Unlike the Markov game where both players have equal positions in the optimization problem~\citep{littman1994markov}, in our Stackelberg game formulation, the leader and follower strive to resolve the following optimization issues:
\begin{equation}
\begin{aligned}
&\min\limits_{\pi_{av}\in \Pi_{av}}\left\{\mathcal{L}_{\pi_{av}}(\pi_{av},\pi_{bv})\right|\pi_{bv}\in\arg\min\limits_{y\in \Pi_{bv}}\mathcal{L}_{\pi_{bv}}(\pi_{av},y)\} \\
&\min\limits_{\pi_{bv}\in \Pi_{bv}}\mathcal{L}_{\pi_{bv}}(\pi_{av},\pi_{bv})
\end{aligned}
\end{equation}
Here, $\mathcal{L}_{\pi_{av}}$ and $\mathcal{L}_{\pi_{bv}}$ represent the loss functions of AV and BV policy networks, respectively. Specifically, they take the following forms (here we use the formula in SAC~\citep{SAC}):
\begin{equation}
\begin{aligned}
&\mathcal{L}_{\pi_{av}}=\alpha\log\pi_{av}(\mathbf{a}_{t,av}|\mathbf{s}_{t})-\operatorname*{min}_{j=1,2}Q_{\text{targ},j}^{av}(\mathbf{s}_t,\mathbf{a}_{t,av},\mathbf{a}_{t,bv})\\
&\mathcal{L}_{\pi_{bv}}=\alpha\log\pi_{bv}(\mathbf{a}_{t,bv}|\mathbf{s}_{t})-\operatorname*{min}_{j=1,2}Q_{\text{targ},j}^{bv}(\mathbf{s}_t,\mathbf{a}_{t,bv},\mathbf{a}_{t,av})
\end{aligned}
\end{equation}
The key challenge in Stackelberg game is how to solve the $\arg\min$ problem as the solution of $\pi_{bv}$ still incorporates the gradient of $\pi_{av}$. To this end, we employ the gradient descent method~\citep{implicit_ld_in_sg} to find the equilibrium where the leader uses its total derivative:
\begin{equation}
\begin{aligned}
\nabla_{\pi_{av}}\mathcal{L}_{\pi_{av}}(\pi_{av},\mathcal{F}(\pi_{av}))=
\left(\nabla_{\pi_{av}}\mathcal{L}_{\pi_{av}}(\pi)-\nabla_{\pi_{av}}\nabla_{\pi_{bv}}\mathcal{L}_{\pi_{bv}}(\pi)\left[\nabla_{\pi_{bv}}^2\mathcal{L}_{\pi_{bv}}(\pi)\right]^{-1}\nabla_{\pi_{bv}}\mathcal{L}_{\pi_{av}}(\pi)\right)
\end{aligned}
\end{equation}
and the follower responds with $\mathcal{F}:\Pi_{av} \rightarrow \Pi_{bv}$ and applies its own policy update. 
\subsection{Practical Implementation}
\paragraph{Agressiveness Regularization of BV.}\label{ar}
During the update process of BV, we incorporate agressiveness regularization mechanism to penalize BV for generating scenarios that are excessively challenging for the current AV. 
This is because the optimization objective of BV policy does not have any constraints related to AV, making it prone to exploring highly aggressive scenarios, leading to game imbalance.
Specifically, this manifests in the form of a high variance between gradients during back propagation. Therefore, it is necessary to incorporate aggressiveness regularization over the optimization of the policy network of BV.

Borrowing the loss functions for the policy network in Soft Actor-Critic (SAC)~\citep{SAC}, we define the loss function of BV Critic with the added regularization as follows:
\begin{equation}
\mathcal{L}_{\pi_{bv}}=\alpha\log\pi_{bv}(\mathbf{a}_{t,bv}|\mathbf{s}_{t})-\operatorname*{min}_{j=1,2}Q_{\text{targ},j}^{bv}(\mathbf{s}_t,\mathbf{a}_{t,bv},\mathbf{a}_{t,av}) - \beta\cdot \operatorname*{min}_{j=1,2}Q_{\text{targ},j}^{av}(\mathbf{s}_t,\mathbf{a}_{t,av},\mathbf{a}_{t,bv})
\end{equation}\label{targetvalue}
Here, $\beta\cdot \operatorname*{min}_{j=1,2}Q_{\text{targ},j}^{av}(\mathbf{s}_t,\mathbf{a}_{t,av},\mathbf{a}_{t,bv})$ is the regularization term, where $\beta \in [0,+\infty)$ determines the degree of regularization. By adding $Q_{\text{targ}}^{av}$ into the loss of BV policy, this regularization restricts BV from generating scenarios too difficult for AV to solve, enabling the robust improvement of AV. We present the results of ablation experiments on regularization term in Sec.~\ref{IV-B.3} Ablation Results.
\paragraph{Update Frequency.}\label{uf}
Apart from update method, update frequency of AV and BV also plays a significant role in training results. It is worth considering the potential consequences when the follower increases the difficulty of scenarios too frequently or the leader's progress is relatively slow. In such cases, the optimization trajectory of AV in the objective function space may not have had the opportunity to advance, as frequent updates of BV have already resulted in significant alterations to the objective function space structure. Consequently, this can result in a divergence in the game process. Therefore, the update frequency ratio between AV and BV ${f_{av}}/{f_{bv}}$ should be higher than a certain lower bound. We present the results of ablation experiments on update frequency in Sec.~\ref{IV-B.3} Ablation Results.
\paragraph{Pretraining of AV.}
Before the game begins, we pretrain AV to convergence in SUMO. This allows AV policy to have reasonable constraints at the start of the game, ensuring stable gameplay.

\section{Evaluation}\label{eval}
In this section, we provide experimental evidence to demonstrate the effectiveness of our modeling approach in optimizing autonomous driving policies. We mainly address the following questions to support our claims:

1) Can SDM AV and BV outperform other baselines in various aspects (safety for AV, quality and risky level for BV)? (Sec.~\ref{IV-B.1} Overall Comparative Experiments)

2) Can SDM AV and BV both \textbf{continually} achieve better performance during the game? (Sec.~\ref{IV-B.2} Controlled Experiments)

\subsection{Experimental Settings}
\paragraph{Environment Setup.}
In this work, we import naturalistic vehicle trajectories data from HighD dataset~\citep{HighD} as initial state $\mathbf{s}_0$ in SUMO simulator~\citep{behrisch2011sumo}, in which we set up an environment that corresponds with the three-line highway in HighD for scenario replay.
We then select segments from higher dimensional and lowest dimensional scenarios that contain 6 and 2 vehicles respectively in HighD, arbitrarily choosing one of them being AV and the others being BVs. 
The processed data points are standardized as transitions $(\mathbf{s}, \mathbf{a}, r, \mathbf{s}', \text{done})$ that can be leveraged in downstream RL algorithms directly.

\paragraph{Evaluation Metrics.}
To evaluate the performance of AV driving policy and the quality of BV-generated scenarios, we select widely recognized metrics in autonomous driving.
AV \textbf{Collision Rate (CR)} and BV CR present the percentage of AV-BV and BV-BV collision scenarios out of all evaluation scenarios respectively.
Additionally, in order to measure the temporal and spatial density of AV-BV collision occurrence, 
we adopt \textbf{Average Collision Frequency Per Second (CPS)} and \textbf{Average Collision Frequency Per 100 Meter (CPM)}~\citep{re2h2o}, given by AV-BV collision numbers $N_{col}$ averaged by total testing time $T_{tt}$ and total AV testing distance $D_{tt}$ respectively: $CPS = N_{col}/T_{tt}, CPM = N_{col}/D_{tt}$.

\paragraph{Testing AV/BV Driving Policies.}
Both AV and BV are tested using various policy models.
For RL AV agent in game (RL-AV), we apply 2 BV driving policies for testing: 
1) SUMO Car-Following Model (\textbf{SUMO-BV})~\citep{SUMO-follow} with SUMO lane-changing model~\citep{SUMO-lane};
2) RL BV agent in game (\textbf{RL-BV}).
For RL BV agent in game (RL-BV), we also apply 2 AV driving polices for testing:
1) RL AV agent pretrained in SUMO-BV environment (\textbf{pretrained-AV}), which is a fixed policy;
2) RL AV agent in game (\textbf{RL-AV}).
We conduct three sets of experiments: pretrained-AV vs RL-BV, RL-AV vs SUMO-BV, RL-AV vs RL-BV. 
It is worth noting that the RL-AV and RL-BV tested against each other are both obtained within the \textbf{same} game setting.

\begin{wrapfigure}{r}{0.6\linewidth}

    \vspace{-15pt}
    \subfloat{
        \centering
        \includegraphics[width=0.49\linewidth]{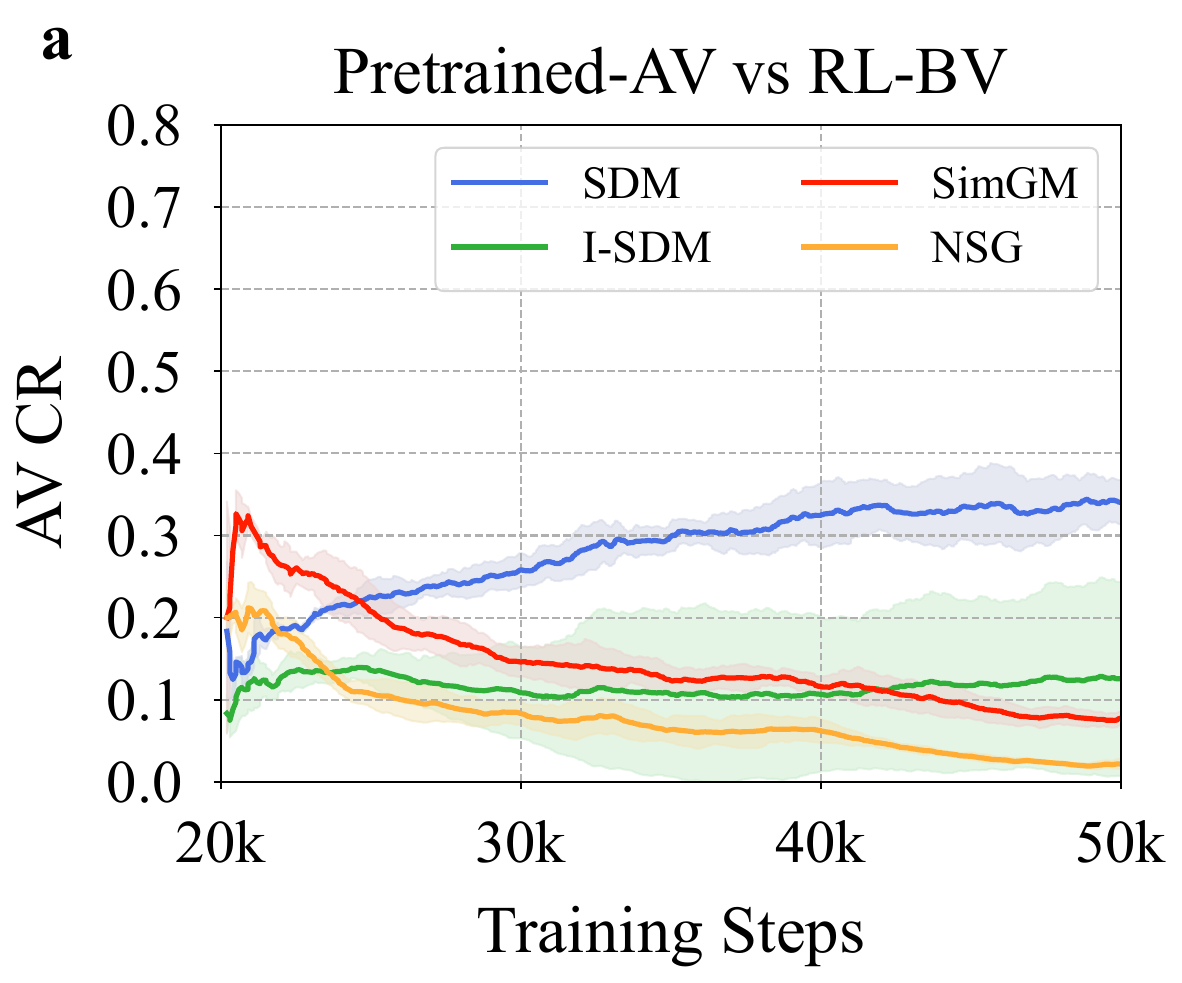}
        \label{ca}
    }
    \subfloat{
        \centering
        \includegraphics[width=0.49\linewidth]{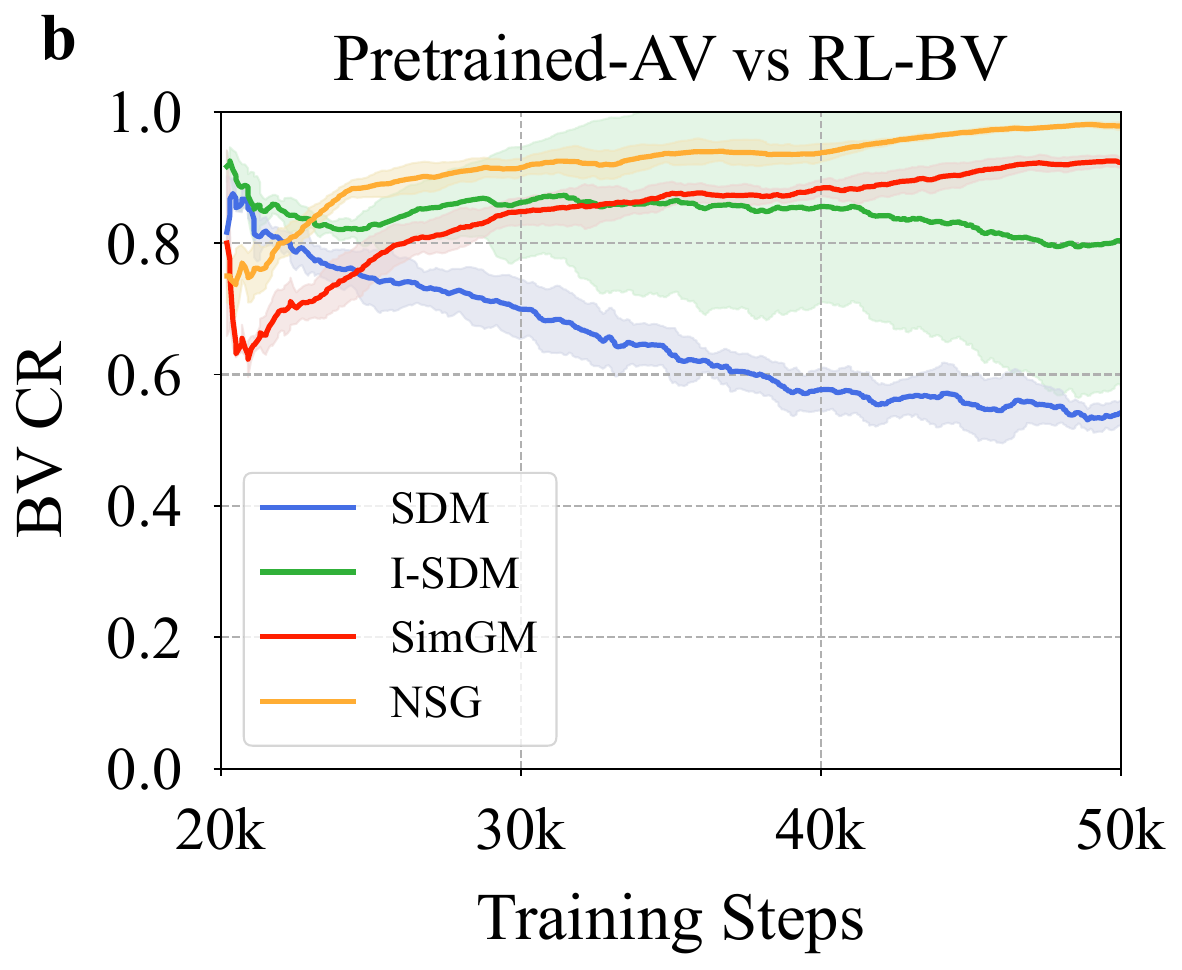}
        \label{cb}
    }
    
    \subfloat{
        \centering
        \includegraphics[width=0.49\linewidth]{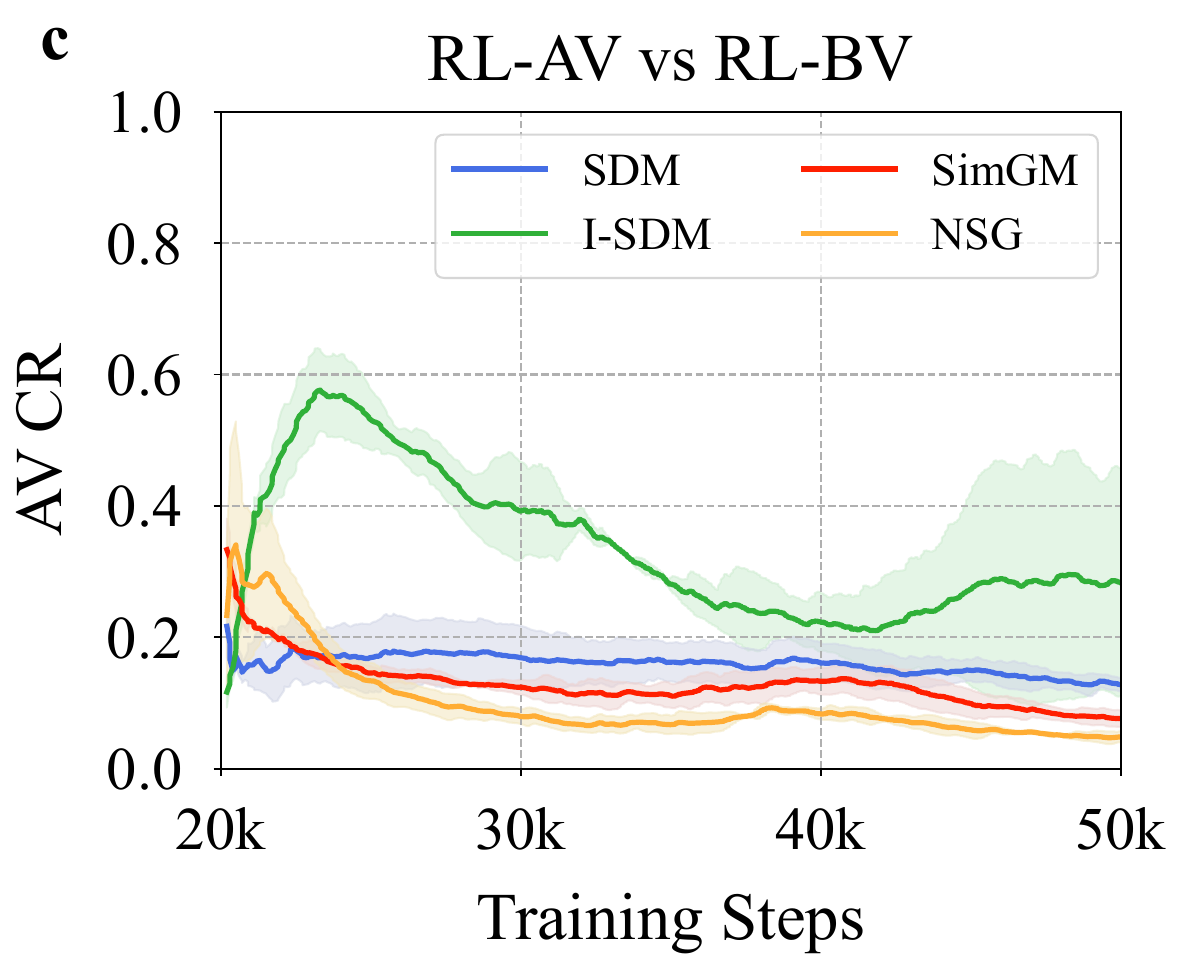}
        \label{cc}
    }    
    \subfloat{
        \centering
        \includegraphics[width=0.49\linewidth]{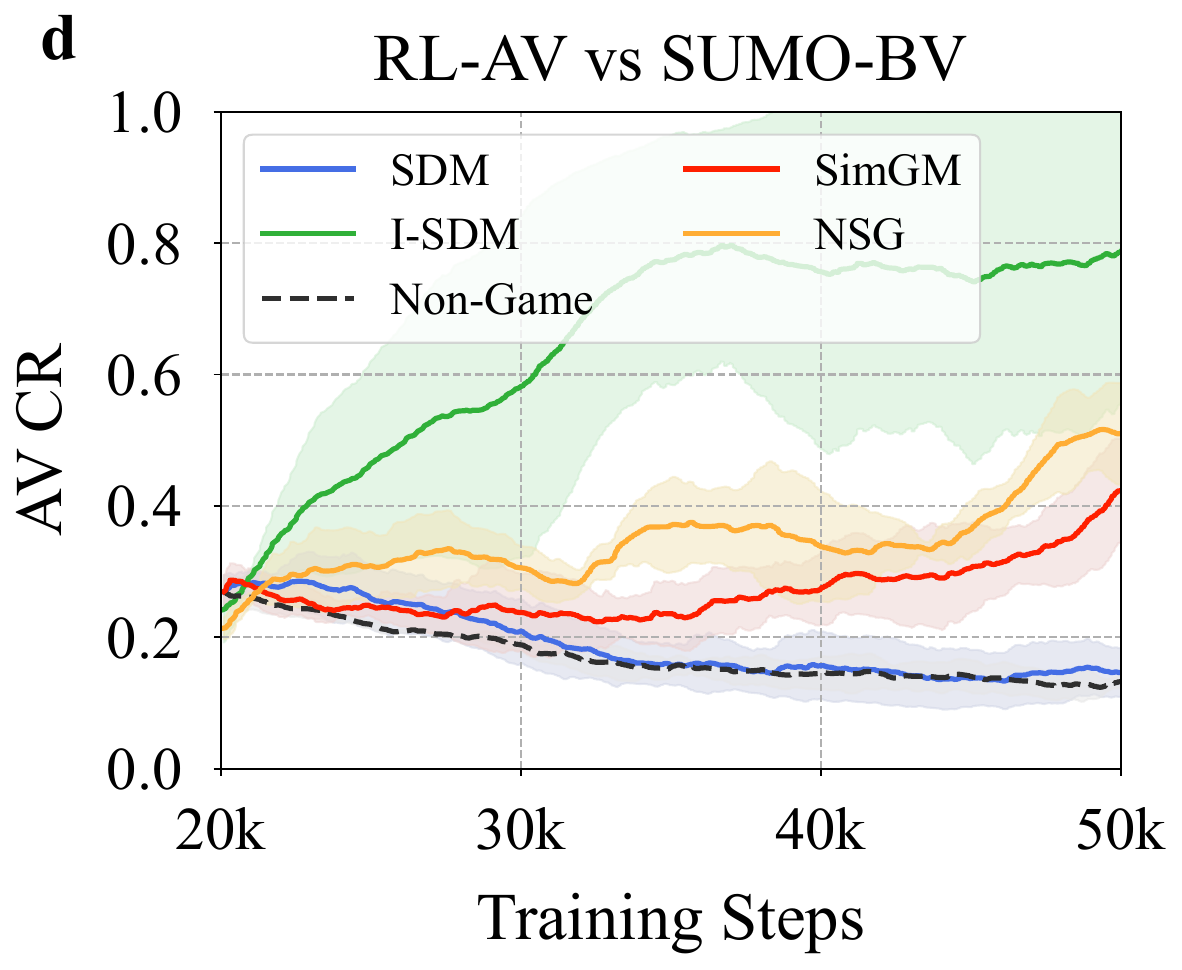}
        \label{cd}
    }
    \caption{Comparison between AV, BV agents of \textbf{SDM}, \textbf{SimGM}, \textbf{I-SDM} and \textbf{NSG}. All scenarios contain 6 vehicles. AVs of SDM and all baselines are pretrained to convergence before 20k step mark. The ablation parameters for SDM are $\beta=0.2,{f_{av}}/{f_{bv}}={5}/{1}$. The subfigures show CRs of \textbf{(a)} pretrained-AV, \textbf{(b)} RL-BV, \textbf{(c)} RL-AV and \textbf{(d)} RL-AV when tested with \textbf{(a)} RL-BV, \textbf{(b)} pretrained-AV, \textbf{(c)} RL-BV and \textbf{(d)} SUMO-BV respectively (all results are computed using exponential smoothing with a 0.99 coefficient and averaged over 3 random seeds).} 
    \vspace{-40pt}

\end{wrapfigure}

\paragraph{Baselines.}
We design four baselines for comparisons:

1) \textbf{Non-Game}: we train RL-AV in SUMO-BV traffic flow with no game modeling.

2) \textbf{Simultaneous Game Model (SimGM)}: we train AV to operate in the presence of destabilizing adversary BVs that apply disturbance forces to the system, which is a zero-sum simultaneous game (Robust Adversarial Reinforcement Learning, RARL)~\citep{RARL}. 

3) \textbf{Naive Sequential Game (NSG)}: we employ the SAC~\citep{SAC} algorithm to adversarially train AV and BV models, where AV and BV take turns acting as the \textbf{Agent} and \textbf{Environment} sequentially.

4) \textbf{Inversed Stackelberg Driver Model (I-SDM)}: we set AV as follower and BV as leader, in contrast to the original SDM model, with $\beta=0$ and ${f_{av}}/{f_{bv}}={1}/{1}$.

\subsection{Experimental Results}

\paragraph{Overall Comparative Experiments.}\label{IV-B.1}

In this paragraph, we analyze the training and testing performance of SDM against other baselines and provide a rigorous argument about the superiority of SDM. 
We initiate the state information of five BVs and one AV (6-vehicle scenario) from the processed six-vehicle scenarios from HighD. Before SDM training, we conduct pretraining on the AV model with SAC and set BV to ``SUMO-BV''. Subsequently, both AV and BV are set as RL agents trained with SDM during the game.
\newtheorem{conclusion}{Conclusion}
\begin{conclusion}\label{con-1}
    \textbf{The BV of SDM outperforms other baselines in both quality and risky level.}
\end{conclusion}

Fig.~\ref{ca} and~\ref{cb} illustrate the variation in performance of BVs of different game settings when tested in pretrained-AV. 
A higher pretrained-AV CR (Fig.~\ref{ca}) implies that the SDM BV model is more risky. The ``BV CR'' (BV-BV collision rate) in Fig.~\ref{cb} represents the quality of BV, where SDM BV exhibits highest quality with lowest BV CR. 
Tab.~\ref{tab:1} shows quantitative evidence in ``pretrained-AV vs RL-BV'', where AV CR, CPS and CPM in SDM is the highest and BV CR in SDM is the lowest compared to other baselines.

It can be observed that SDM makes BV increasingly difficult (Fig.~\ref{ca}), while the collision rate between BVs (BV CR) decreases (Fig.~\ref{cb}), exhibiting overwhelming advantages. 
The comparison between SDM and I-SDM (blue line and green line in Fig.~\ref{ca} and~\ref{cb}) indicates that modeling BV as the follower allows BV to explore more risky actions for AV (higher pretrained-AV CR) while ensuring its own quality (lower RL-BV CR).

\begin{conclusion}\label{con-2}
    \textbf{The AV of SDM outperforms other baselines in safety.}
\end{conclusion}

Fig.~\ref{cc} and~\ref{cd} depicts the dynamic evolution of AV performance in different game settings when tesed in RL-BV and SUMO-BV. Quantitative results are shown in ``RL-AV vs RL-BV'' and ``RL-AV vs SUMO-BV'' in Tab.~\ref{tab:1}.

In RL-BV testing environment (Fig.~\ref{cc} and Tab.~\ref{tab:1} ``RL-AV vs RL-BV''), the results are consistent with Con.~\ref{con-1}. Despite SimGM and NSG showing lower AV CR, it is primarily due to their BV CR (BV-BV collision rate) approaching $100\%$ (Fig.~\ref{cb}), meaning each trajectory ends prematurely before AV has any chance to collide. This results in high CPS. In contrast, AV of SDM continues to exhibit exceptional performance even challenged by aggressive and high-quality BVs (Con.~\ref{con-1}), demonstrating that SDM significantly enhances AV capability to handle risky scenarios.

In SUMO-BV testing environment (Fig.~\ref{cd} and Tab.~\ref{tab:1} ``RL-AV vs SUMO-BV''), SDM AV is slightly inferior to Non-Game AV (higher AV CR), primarily due to the fact that the former is trained in continually challenging RL-BV, while the latter is trained in SUMO. Testing SDM AV in simple SUMO-BV tends to yield more conservative results.

By comparing the performance of pretrained-AV and SDM RL-AV under SDM RL-BV environment (blue lines of Fig.~\ref{ca} and~\ref{cc}), it becomes evident that the pretrained-AV is unable to effectively handle the encountered RL-BV of SDM (higher AV CR in Fig.~\ref{ca}). However, the RL-AV of SDM demonstrate a rising ability to cope with the gradually increased RL-BV (lower AV CR in Fig.~\ref{cc}), indicating that the SDM strengthens AV capabilities and enables it to adapt and overcome the challenges presented by RL-BV. Furthermore, SDM AV exhibits an overwhelming advantage over I-SDM (lower RL-AV CR in both RL-BV and SUMO-BV environment), underscoring the advantage of modeling AV as the leader in dealing with challenging scenarios generated by BVs.


\begin{wrapfigure}{r}{0.5\linewidth}
    \vspace{-10pt}
    \includegraphics[width=1\linewidth]{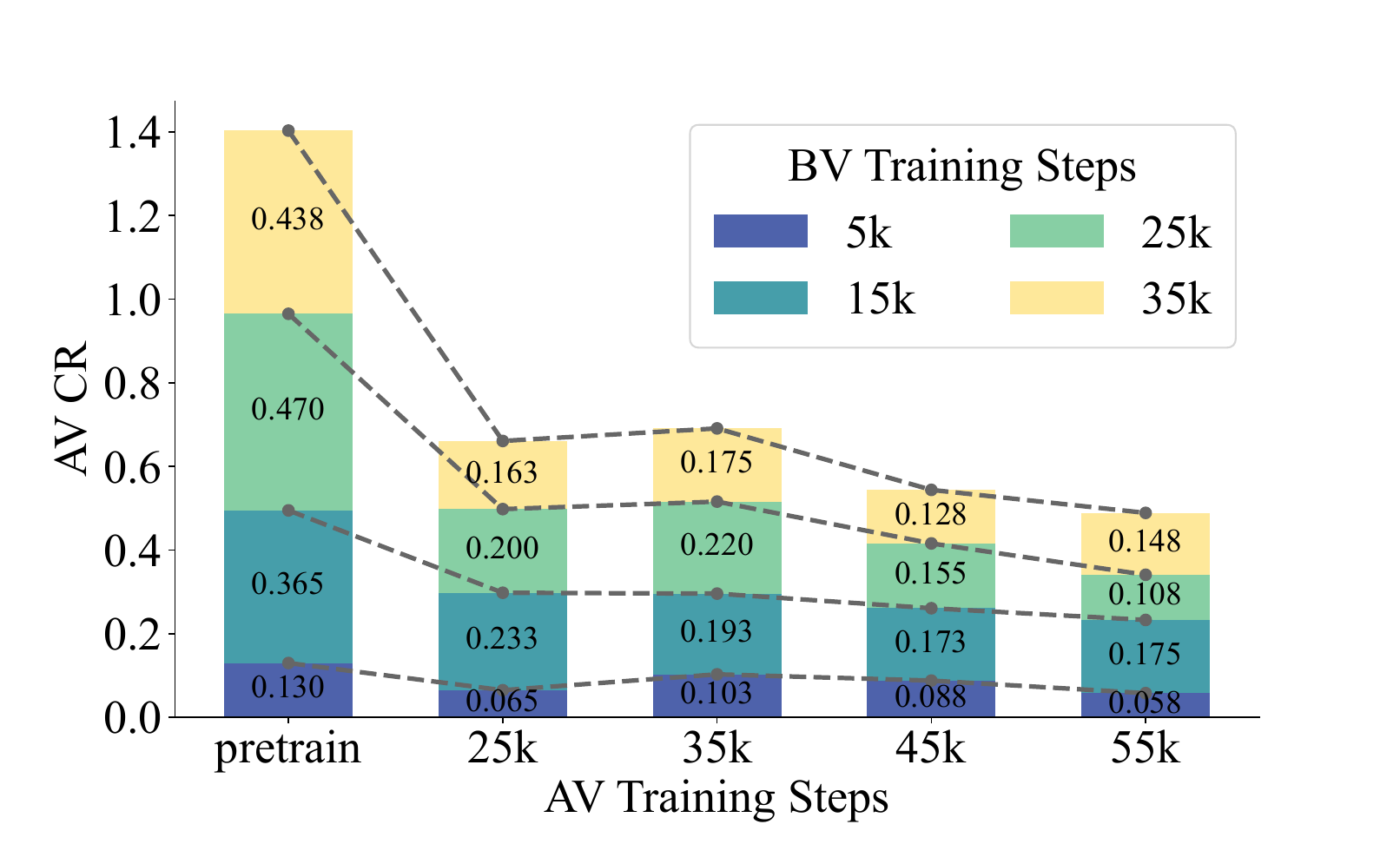}
    \caption{Controlled experiment results of AV agents in varying training durations against BV agents (5 BVs) of differential difficulty levels.}
    \vspace{-10pt}
    \label{fig:3}
\end{wrapfigure}

\paragraph{Controlled Experiments.}\label{IV-B.2}

In this paragraph, experiments are conducted \textbf{in 6-vehicle scenarios} utilizing SDM BV of differential difficulty levels to evaluate AV at various training stages, including pretraining (20k) and SDM training (25k, 35k, 45k, 55k steps). It is worth noting that BV training begins after AV pretraining phase (20k). The resultant AV CR, as illustrated in Fig.~\ref{fig:3}, serves to exhibit the efficacy of SDM.

For each BV model, under the same AV training conditions, as the number of BV training steps increased (different colors), a continuous ascent in AV CR was observed. This signifies that over the course of game training, the difficulty level of BV continuously intensified, enabling the generation of increasingly challenging scenarios from the BVs.

For each AV model, at the same difficulty level of BV (same color), AV collision rate remained elevated during the game (illustrated by the dashed line), 
indicating progressive policy enhancement of AV. 

\begin{table*}[t]
\centering
\renewcommand{\arraystretch}{1.2}
\setlength\tabcolsep{5.5pt} 
    \caption{\textbf{Assessment on comparisons with other baselines and ablations.} Results are averaged over 3 random seeds. The optimal values and second-best values for each column in both 2-vechicle and 6-vehicle scenarios are marked in \textcolor{red}{red} and \textcolor{blue}{blue} respectively. Two ablation parameters for SDM represent aggressive regularization scale $\beta=0.00,0.20,1.00,2.00,10.00$ and AV/BV update frequency ratio $\frac{f_{av}}{f_{bv}}=\frac{5}{1},\frac{1}{1},\frac{1}{5}$ successively.}
    \resizebox{\textwidth}{!}{
    \begin{tabular}{ccccccccccccccc}
        \toprule
        \multicolumn{3}{c}{\multirow{2}{*}{AV-BV Modeling}}& \multicolumn{4}{c}{pretrained-AV vs RL-BV} & \multicolumn{4}{c}{RL-AV vs SUMO-BV} & \multicolumn{4}{c}{RL-AV vs RL-BV}\\
        \cmidrule(r){4-7} \cmidrule(r){8-11} \cmidrule(r){12-15}
        &&& AV CR$\uparrow$ & BV CR$\downarrow$ & CPS$\uparrow$ & CPM$\uparrow$ & AV CR$\downarrow$ & BV CR$\uparrow$ & CPS$\downarrow$ & CPM$\downarrow$ & AV CR$\downarrow$ & BV CR$\downarrow$ & CPS$\downarrow$ & CPM$\downarrow$ \\
        \midrule
        \multirow{11}{*}{\rotatebox{90}{5 BVs}}
        &\multirow{4}{*}{\rotatebox{90}{Baselines}}
        & Non-Game & / & / & / & / & \textcolor{blue}{12.4\%} & 0.667\% & \textcolor{blue}{0.0368} & \textcolor{red}{0.0830} & / & / & / & /\\
        && SimGM & 6.89\% & 93.5\% & \textcolor{red}{0.133} & 0.116 & 57.0\% & 0.758\% & 0.225 & 0.582 & \textcolor{blue}{6.45\%} & 93.6\% & 0.138 & 0.112 \\
        && NSG & 1.93\% & 98.1\% & 0.0338 & 0.0334 & 61.5\% & 0.833\% & 0.305 & 0.684 & \textcolor{red}{4.87\%} & 95.1\% & 0.0868 & \textcolor{red}{0.0820} \\
        && I-SDM & 16.5\% & 72.9\% & 0.0883 & 0.146 & 80.3\% & 0.687\% & 0.565 & 1.04 & 24.9\% & 71.2\% & 0.250 & 0.316 \\
        \cmidrule{2-15}
        &\multirow{7}{*}{\rotatebox{90}{    -SDM}}
        & 0.20, 1:5 & 31.9\% & 58.8\% & 0.115 & 0.252 & 17.2\% & 0.833\% & 0.0495 & 0.111 & 13.0\% & 68.9\% & 0.0453 & 0.114 \\
        && 0.20, 1:1 & 34.1\% & 57.0\% & 0.128 & 0.277 & 17.9\% & 0.909\% & 0.0508 & 0.140 & 13.3\% & 73.0\% & 0.0458 & 0.117 \\
        && 0.20, 5:1 & 33.2\% & \textcolor{blue}{56.4\%} & 0.125 & 0.269 & 15.7\% & 0.909\% & 0.0433 & 0.123 & 13.4\% & 68.6\% & \textcolor{blue}{0.0435} & 0.112 \\
        && 0.00, 5:1 & 32.3\% & 58.9\% & 0.123 & 0.266 & 16.1\% & \textcolor{blue}{1.02\%} & 0.0453 & 0.128 & 13.3\% & 68.9\% & 0.0463 & 0.118 \\
        \rowcolor{gray!20}\cellcolor{white}&\cellcolor{white}& 1.00, 5:1 & \textcolor{red}{36.3\%} & \textcolor{red}{51.2\%} & \textcolor{blue}{0.131} & \textcolor{red}{0.287} & 18.4\% & \textcolor{red}{1.17\%} & 0.0513 & 0.145 & 12.3\% & \textcolor{red}{65.2\%} & \textcolor{red}{0.0403} & 0.105 \\
        && 2.00, 5:1 & \textcolor{blue}{34.2\%} & 56.7\% & 0.128 & \textcolor{blue}{0.284} & 14.4\% & 0.441\% & 0.0397 & 0.102 & 15.7\% & \textcolor{blue}{66.6\%} & 0.0534 & 0.139 \\
        && 10.00, 5:1 & 18.9\% & 80.1\% & 0.0990 & 0.191 & \textcolor{red}{11.5\%} & 0.694\% & \textcolor{red}{0.0317} & \textcolor{blue}{0.0907} & 9.32\% & 90.7\% & 0.0454 & \textcolor{blue}{0.0979} \\
        \midrule
        \midrule
        \multirow{11}{*}{\rotatebox{90}{1 BV}}
        &\multirow{4}{*}{\rotatebox{90}{Baselines}}
        & Non-Game & / & / & / & / & \textcolor{red}{3.40\%} & 0.00\% & \textcolor{red}{0.00893} & \textcolor{red}{0.0247} & / & / & /  & /\\
        && SimGM & 28.9\% & 71.1\% & \textcolor{red}{0.586} & \textcolor{blue}{1.39} & 56.8\% & 0.00\% & 0.203 & 0.674 & 24.0\% & 76.2\% & 0.481 & 1.18 \\
        && NSG & 50.3\% & 39.9\% & \textcolor{blue}{0.483} & \textcolor{red}{1.51} & 41.0\% & 0.00\% & 0.138 & 0.491 & 49.8\% & 40.7\% & 0.457 & 1.58 \\
        && I-SDM & 13.9\% & 85.5\% & 0.188 & 0.536 & 98.0\% & 0.00\% & 1.13 & 3.50 & 40.4\% & 66.4\% & 0.596 & 1.64 \\
        \cmidrule(r){2-15}
        &\multirow{7}{*}{\rotatebox{90}{SDM}}
        & 0.20, 1:5 & \textcolor{blue}{53.6\%} & 42.0\% & 0.443 & 1.25 & 16.9\% & 0.00\% & 0.0486 & 0.172 & 51.2\% & 44.3\% & 0.443 & 1.31 \\ 
        && 0.20, 1:1 & \textcolor{red}{54.0\%} & 37.0\% & 0.376 & 1.07 & 14.3\% & 0.00\% & 0.0400 & 0.141 & 50.4\% & 39.4\% & 0.361 & 1.13 \\ 
        && 0.20, 5:1 & 19.9\% & 45.7\% & 0.0656 & 0.191 & 10.3\% & 0.00\% & \textcolor{blue}{0.0282} & 0.0912 & \textcolor{red}{6.73\%} & 54.4\% & \textcolor{red}{0.0220} & \textcolor{red}{0.0752} \\ 
        && 0.00, 5:1 & 15.4\% & 73.8\% & 0.0714 & 0.212 & 10.8\% & 0.00\% & 0.0294 & 0.0942 & \textcolor{blue}{10.1\%} & 74.9\% & 0.0456 & 0.152 \\ 
        && 1.00, 5:1 & 35.7\% & \textcolor{red}{16.3\%} & 0.113 & 0.334 & 11.8\% & 0.00\% & 0.0317 & 0.109 & 18.2\% & \textcolor{blue}{17.5\%} & 0.0554 & 0.332 \\ 
        \rowcolor{gray!20}\cellcolor{white}&\cellcolor{white}& 2.00, 5:1 & 33.6\% & \textcolor{blue}{19.2\%} & 0.112 & 0.323 & \textcolor{blue}{8.98\%} & 0.00\% & \textcolor{blue}{0.0282} & \textcolor{blue}{0.0910} & 14.1\% & \textcolor{red}{16.2\%} & \textcolor{blue}{0.0412} & \textcolor{blue}{0.144} \\ 
        && 10.00, 5:1 & 29.3\% & 49.4\% & 0.122 & 0.361 & 12.8\% & 0.00\% & 0.0351 & 0.122& 18.6\% & 49.0\% & 0.0678 & 0.245 \\ 
        \bottomrule
    \end{tabular}
    }
    \vspace{-4mm}
    \label{tab:1}
\end{table*}

\paragraph{Ablation Results.}\label{IV-B.3}
Tab.~\ref{tab:1} showcases the results of ablation experiments in 6-vehicle scenarios and 2-vehicle scenarios on aggressiveness regularization scale $\beta=0.00,0.20,1.00,2.00,10.00$ and update frequency ratio ${f_{av}}/{f_{bv}}={5}/{1},{1}/{1},{1}/{5}$. 

For aggressiveness regularization, \textbf{in 6-vehicle scenarios}, it can be observed that $\beta=1.00$ results in riskier and higher-quality BVs, reflected in high AV CPS and low BV CR in Pretrained-AV vs RL-BV. Also, it leads to better performance in AV, reflected in low AV CPS and low AV CPM in RL-AV vs RL-BV.
An excessively large $\beta=10$ yields a disproportionately small BV reward share in Eq.~\ref{targetvalue}, resulting in lower-quality (high BV CR in pretrained-AV vs RL-BV) and less risky (low AV CR in pretrained-AV vs RL-BV) BV. Conversely, an overly small $\beta=0$ permits BVs to explore excessively challenging scenarios, limiting AV performance (high AV CR in RL-AV vs RL-BV).
A similar pattern also emerges \textbf{in 2-vehicle scenarios}, where $\beta=2.00$ shows relatively higher-quality (lower BV CR in pretrained-AV vs RL-BV) and riskier (higher AV CR in pretrained-AV vs RL-BV) BV under the same ${f_{av}}/{f_{bv}}=5/1$.
It is evident that the regularization term constrains the exploration of BV from too aggressive behaviors in action space, consistent with Sec.~\ref{ar} ``Agressiveness Regularization of BV''.

In the case of AV/BV update frequency ratio ${f_{av}}/{f_{bv}}$, \textbf{in 6-vehicle scenarios}, a larger ratio leads to the learning of higher-quality BVs, as evident in the lower BV CR in Pretrained-AV vs RL-BV and RL-AV vs RL-BV. Additionally, higher ${f_{av}}/{f_{bv}}$ results in better AV performance, as reflected in lower AV CPS and CPM in both RL-AV vs SUMO-BV and RL-AV vs RL-BV. 
However, \textbf{in 2-vehicle scenarios}, a larger ${f_{av}}/{f_{bv}}={5}/{1}$ leads to the least challenging (lowest AV CR in pretrained-AV vs RL-BV) BV, while resulting in the safest (lowest AV CR in RL-AV vs RL-BV) AV. This implies that a lower $f_{bv}$ can maintain a relatively stable structure of the AV's objective function, giving AV sufficient time to update, while higher $f_{bv}$ allows better BV update, which aligns with the analysis in Sec.~\ref{uf} ``Update Frequency''.

\paragraph{Experiments on Scenarios with Different Dimensions.}
As shown in Tab.~\ref{tab:1}, it reveals that NSG achieves competitive BV performance (high AV CR and low BV CR in pretrained-AV vs RL-BV) in 2-vehicle scenarios. However, in the higher dimensional 6-vehicle scenarios, the baselines BVs exhibit a sharp decline in performance, while SDM BV shows little. This highlights the superiority of SDM, especially as the scenario dimension increases.

\section{Conclusion and Future Work}\label{conclusion}
In this paper, we propose a novel closed-loop scenario-based autonomous driving framework integrating scenario generation, AV testing and AV improvement.
The hierarchical nature of AV-BV interaction is ingeniously characterized in a sequential Stackelberg game paradigm, referred to as the Stackelberg Driver Model (SDM). 
SDM prioritizes AV by informing it that BV would respond with the best strategy to challenge AV at each training iteration.
Consequently, AV continually acquires strategies to navigate through riskier situations, while BV develops the capacity to generate progressively challenging scenarios for AV.
Compared to other baselines, SDM demonstrates overwhelming advantages in both AV and BV, capable of yielding high-performance AV in both regular and risky scenarios, as well as high-quality (low BV CR) and challenging (high AV CR) BV. 
Our ablations stress and justify the hierarchical relationship between AV and BVs and its consistency with the inherent nature of leader-follower game process. 
In the future, we aspire to conduct experiments with more intricate road topologies, as well as apply SDM to other complex intelligent systems that also suffer from long-tail effects.
\bibliographystyle{named}
\bibliography{mylib}

\begin{thebibliography}{}

\bibitem[\protect\citeauthoryear{Abeysirigoonawardena \bgroup \em et al.\egroup }{2019}]{Generating_adversarial_Scenarios}
Yasasa Abeysirigoonawardena, Florian Shkurti, and Gregory Dudek.
\newblock Generating adversarial driving scenarios in high-fidelity simulators.
\newblock In {\em 2019 International Conference on Robotics and Automation (ICRA)}, pages 8271--8277, 2019.

\bibitem[\protect\citeauthoryear{Behrisch \bgroup \em et al.\egroup }{2011}]{behrisch2011sumo}
Michael Behrisch, Laura Bieker, Jakob Erdmann, and Daniel Krajzewicz.
\newblock Sumo--simulation of urban mobility: an overview.
\newblock In {\em Proceedings of SIMUL 2011, The Third International Conference on Advances in System Simulation}. ThinkMind, 2011.

\bibitem[\protect\citeauthoryear{Brocas \bgroup \em et al.\egroup }{2018}]{SeqAndSim}
Isabelle Brocas, Juan~D. Carrillo, and Ashish Sachdeva.
\newblock The path to equilibrium in sequential and simultaneous games: A mousetracking study.
\newblock {\em Journal of Economic Theory}, 178:246--274, 2018.

\bibitem[\protect\citeauthoryear{Chen \bgroup \em et al.\egroup }{2018}]{chen2018model}
Yize Chen, Yishen Wang, Daniel Kirschen, and Baosen Zhang.
\newblock Model-free renewable scenario generation using generative adversarial networks.
\newblock {\em IEEE Transactions on Power Systems}, 33(3):3265--3275, 2018.

\bibitem[\protect\citeauthoryear{Cui \bgroup \em et al.\egroup }{2019}]{cui2019class}
Yin Cui, Menglin Jia, Tsung-Yi Lin, Yang Song, and Serge Belongie.
\newblock Class-balanced loss based on effective number of samples.
\newblock In {\em Proceedings of the IEEE/CVF conference on computer vision and pattern recognition}, pages 9268--9277, 2019.

\bibitem[\protect\citeauthoryear{Ding \bgroup \em et al.\egroup }{2020}]{ding2020learning}
Wenhao Ding, Baiming Chen, Minjun Xu, and Ding Zhao.
\newblock Learning to collide: An adaptive safety-critical scenarios generating method.
\newblock In {\em 2020 IEEE/RSJ International Conference on Intelligent Robots and Systems (IROS)}, pages 2243--2250. IEEE, 2020.

\bibitem[\protect\citeauthoryear{Ding \bgroup \em et al.\egroup }{2023}]{ding2023survey}
Wenhao Ding, Chejian Xu, Mansur Arief, Haohong Lin, Bo~Li, and Ding Zhao.
\newblock A survey on safety-critical driving scenario generation—a methodological perspective.
\newblock {\em IEEE Transactions on Intelligent Transportation Systems}, 2023.

\bibitem[\protect\citeauthoryear{Erdmann}{2015}]{SUMO-lane}
Jakob Erdmann.
\newblock Sumo's lane-changing model.
\newblock In Michael Behrisch and Melanie Weber, editors, {\em Modeling Mobility with Open Data}, pages 105--123, Cham, 2015. Springer International Publishing.

\bibitem[\protect\citeauthoryear{Feng \bgroup \em et al.\egroup }{2021}]{feng2021intelligent}
Shuo Feng, Xintao Yan, Haowei Sun, Yiheng Feng, and Henry~X Liu.
\newblock Intelligent driving intelligence test for autonomous vehicles with naturalistic and adversarial environment.
\newblock {\em Nature communications}, 12(1):748, 2021.

\bibitem[\protect\citeauthoryear{Feng \bgroup \em et al.\egroup }{2023}]{DRL}
Shuo Feng, Haowei Sun, Xintao Yan, Haojie Zhu, Zhengxia Zou, Shengyin Shen, and Henry~X Liu.
\newblock Dense reinforcement learning for safety validation of autonomous vehicles.
\newblock {\em Nature}, 615(7953):620--627, 2023.

\bibitem[\protect\citeauthoryear{Fiez \bgroup \em et al.\egroup }{2019}]{Convergence_of_LD_in_SG}
Tanner Fiez, Benjamin Chasnov, and Lillian~J. Ratliff.
\newblock Convergence of learning dynamics in stackelberg games.
\newblock {\em CoRR}, abs/1906.01217, 2019.

\bibitem[\protect\citeauthoryear{Fiez \bgroup \em et al.\egroup }{2020}]{implicit_ld_in_sg}
Tanner Fiez, Benjamin Chasnov, and Lillian Ratliff.
\newblock Implicit learning dynamics in stackelberg games: Equilibria characterization, convergence analysis, and empirical study.
\newblock In Hal~Daumé III and Aarti Singh, editors, {\em Proceedings of the 37th International Conference on Machine Learning}, volume 119 of {\em Proceedings of Machine Learning Research}, pages 3133--3144. PMLR, 13--18 Jul 2020.

\bibitem[\protect\citeauthoryear{Goodfellow \bgroup \em et al.\egroup }{2020}]{GAN}
Ian Goodfellow, Jean Pouget-Abadie, Mehdi Mirza, Bing Xu, David Warde-Farley, Sherjil Ozair, Aaron Courville, and Yoshua Bengio.
\newblock Generative adversarial networks.
\newblock {\em Commun. ACM}, 63(11):139–144, oct 2020.

\bibitem[\protect\citeauthoryear{Haarnoja \bgroup \em et al.\egroup }{2018}]{SAC}
Tuomas Haarnoja, Aurick Zhou, Pieter Abbeel, and Sergey Levine.
\newblock Soft actor-critic: Off-policy maximum entropy deep reinforcement learning with a stochastic actor.
\newblock In {\em International conference on machine learning}, pages 1861--1870. PMLR, 2018.

\bibitem[\protect\citeauthoryear{Hanselmann \bgroup \em et al.\egroup }{2022}]{King}
Niklas Hanselmann, Katrin Renz, Kashyap Chitta, Apratim Bhattacharyya, and Andreas Geiger.
\newblock King: Generating safety-critical driving scenarios for robust imitation via kinematics gradients.
\newblock In {\em European Conference on Computer Vision}, pages 335--352. Springer, 2022.

\bibitem[\protect\citeauthoryear{Holt and Roth}{2004}]{holt2004nash}
Charles~A Holt and Alvin~E Roth.
\newblock The nash equilibrium: A perspective.
\newblock {\em Proceedings of the National Academy of Sciences}, 101(12):3999--4002, 2004.

\bibitem[\protect\citeauthoryear{Huang \bgroup \em et al.\egroup }{2022}]{huang2022robust}
Peide Huang, Mengdi Xu, Fei Fang, and Ding Zhao.
\newblock Robust reinforcement learning as a stackelberg game via adaptively-regularized adversarial training.
\newblock {\em arXiv preprint arXiv:2202.09514}, 2022.

\bibitem[\protect\citeauthoryear{Joel \bgroup \em et al.\egroup }{2002}]{joel2002actor}
Daphna Joel, Yael Niv, and Eytan Ruppin.
\newblock Actor--critic models of the basal ganglia: New anatomical and computational perspectives.
\newblock {\em Neural networks}, 15(4-6):535--547, 2002.

\bibitem[\protect\citeauthoryear{Konda and Tsitsiklis}{1999}]{konda1999actor}
Vijay Konda and John Tsitsiklis.
\newblock Actor-critic algorithms.
\newblock {\em Advances in neural information processing systems}, 12, 1999.

\bibitem[\protect\citeauthoryear{Kou \bgroup \em et al.\egroup }{2008}]{kou2008worst}
Youseok Kou, Huei Peng, and DoHyun Jung.
\newblock Worst-case evaluation for integrated chassis control systems.
\newblock {\em Vehicle System Dynamics}, 46(S1):329--340, 2008.

\bibitem[\protect\citeauthoryear{Krajewski \bgroup \em et al.\egroup }{2018}]{HighD}
Robert Krajewski, Julian Bock, Laurent Kloeker, and Lutz Eckstein.
\newblock The highd dataset: A drone dataset of naturalistic vehicle trajectories on german highways for validation of highly automated driving systems.
\newblock In {\em 2018 21st International Conference on Intelligent Transportation Systems (ITSC)}, pages 2118--2125, 2018.

\bibitem[\protect\citeauthoryear{Kreps}{1989}]{kreps1989nash}
David~M Kreps.
\newblock Nash equilibrium.
\newblock In {\em Game Theory}, pages 167--177. Springer, 1989.

\bibitem[\protect\citeauthoryear{Li and Sethi}{2017}]{stackelberggame}
Tao Li and Suresh~P Sethi.
\newblock A review of dynamic stackelberg game models.
\newblock {\em Discrete \& Continuous Dynamical Systems-B}, 22(1):125, 2017.

\bibitem[\protect\citeauthoryear{Littman}{1994}]{littman1994markov}
Michael~L Littman.
\newblock Markov games as a framework for multi-agent reinforcement learning.
\newblock In {\em Machine learning proceedings 1994}, pages 157--163. Elsevier, 1994.

\bibitem[\protect\citeauthoryear{Liu \bgroup \em et al.\egroup }{2022}]{BLO}
Risheng Liu, Jiaxin Gao, Jin Zhang, Deyu Meng, and Zhouchen Lin.
\newblock Investigating bi-level optimization for learning and vision from a unified perspective: A survey and beyond.
\newblock {\em IEEE Transactions on Pattern Analysis and Machine Intelligence}, 44(12):10045--10067, 2022.

\bibitem[\protect\citeauthoryear{Liu}{1998}]{liu1998stackelberg}
Baoding Liu.
\newblock Stackelberg-nash equilibrium for multilevel programming with multiple followers using genetic algorithms.
\newblock {\em Computers \& Mathematics with Applications}, 36(7):79--89, 1998.

\bibitem[\protect\citeauthoryear{Lowe \bgroup \em et al.\egroup }{2017}]{multiDDPG}
Ryan Lowe, Yi~I Wu, Aviv Tamar, Jean Harb, OpenAI Pieter~Abbeel, and Igor Mordatch.
\newblock Multi-agent actor-critic for mixed cooperative-competitive environments.
\newblock {\em Advances in neural information processing systems}, 30, 2017.

\bibitem[\protect\citeauthoryear{Niu \bgroup \em et al.\egroup }{2021}]{niu2021dr2l}
Haoyi Niu, Jianming Hu, Zheyu Cui, and Yi~Zhang.
\newblock Dr2l: Surfacing corner cases to robustify autonomous driving via domain randomization reinforcement learning.
\newblock In {\em Proceedings of the 5th International Conference on Computer Science and Application Engineering}, pages 1--8, 2021.

\bibitem[\protect\citeauthoryear{Niu \bgroup \em et al.\egroup }{2023}]{re2h2o}
Haoyi Niu, Kun Ren, Yizhou Xu, Ziyuan Yang, Yichen Lin, Yi~Zhang, and Jianming Hu.
\newblock (re)2h2o: Autonomous driving scenario generation via reversely regularized hybrid offline-and-online reinforcement learning.
\newblock In {\em 2023 IEEE Intelligent Vehicles Symposium (IV)}, pages 1--8, 2023.

\bibitem[\protect\citeauthoryear{O'Kelly \bgroup \em et al.\egroup }{2018}]{o2018scalable}
Matthew O'Kelly, Aman Sinha, Hongseok Namkoong, Russ Tedrake, and John~C Duchi.
\newblock Scalable end-to-end autonomous vehicle testing via rare-event simulation.
\newblock {\em Advances in neural information processing systems}, 31, 2018.

\bibitem[\protect\citeauthoryear{Pinto \bgroup \em et al.\egroup }{2017}]{RARL}
Lerrel Pinto, James Davidson, Rahul Sukthankar, and Abhinav~Kumar Gupta.
\newblock Robust adversarial reinforcement learning.
\newblock In {\em International Conference on Machine Learning}, 2017.

\bibitem[\protect\citeauthoryear{Rempe \bgroup \em et al.\egroup }{2022}]{STRIVE}
Davis Rempe, Jonah Philion, Leonidas~J Guibas, Sanja Fidler, and Or~Litany.
\newblock Generating useful accident-prone driving scenarios via a learned traffic prior.
\newblock In {\em Proceedings of the IEEE/CVF Conference on Computer Vision and Pattern Recognition}, pages 17305--17315, 2022.

\bibitem[\protect\citeauthoryear{Song \bgroup \em et al.\egroup }{2014}]{SUMO-follow}
Jie Song, Yi~Wu, Zhexin Xu, and Xiao Lin.
\newblock Research on car-following model based on sumo.
\newblock In {\em The 7th IEEE/International Conference on Advanced Infocomm Technology}, pages 47--55, 2014.

\bibitem[\protect\citeauthoryear{Sun \bgroup \em et al.\egroup }{2021}]{corner}
Haowei Sun, Shuo Feng, Xintao Yan, and Henry~X Liu.
\newblock Corner case generation and analysis for safety assessment of autonomous vehicles.
\newblock {\em Transportation research record}, 2675(11):587--600, 2021.

\bibitem[\protect\citeauthoryear{Von~Stackelberg}{2010}]{von2010market}
Heinrich Von~Stackelberg.
\newblock {\em Market structure and equilibrium}.
\newblock Springer Science \& Business Media, 2010.

\bibitem[\protect\citeauthoryear{Vu \bgroup \em et al.\egroup }{2022}]{spg}
Quoc-Liem Vu, Zane Alumbaugh, Ryan Ching, Quanchen Ding, Arnav Mahajan, Benjamin Chasnov, Sam Burden, and Lillian~J Ratliff.
\newblock Stackelberg policy gradient: evaluating the performance of leaders and followers.
\newblock In {\em ICLR 2022 Workshop on Gamification and Multiagent Solutions}, 2022.

\bibitem[\protect\citeauthoryear{Wachi}{2019}]{wachi2019failure}
Akifumi Wachi.
\newblock Failure-scenario maker for rule-based agent using multi-agent adversarial reinforcement learning and its application to autonomous driving.
\newblock In {\em International Joint Conference on Artificial Intelligence}. International Joint Conferences on Artificial Intelligence, 2019.

\bibitem[\protect\citeauthoryear{Wang \bgroup \em et al.\egroup }{2019}]{wang2019poet}
Rui Wang, Joel Lehman, Jeff Clune, and Kenneth~O Stanley.
\newblock Poet: open-ended coevolution of environments and their optimized solutions.
\newblock In {\em Proceedings of the Genetic and Evolutionary Computation Conference}, pages 142--151, 2019.

\bibitem[\protect\citeauthoryear{Wang \bgroup \em et al.\egroup }{2020}]{wang2020enhanced}
Rui Wang, Joel Lehman, Aditya Rawal, Jiale Zhi, Yulun Li, Jeffrey Clune, and Kenneth Stanley.
\newblock Enhanced poet: Open-ended reinforcement learning through unbounded invention of learning challenges and their solutions.
\newblock In {\em International Conference on Machine Learning}, pages 9940--9951. PMLR, 2020.

\bibitem[\protect\citeauthoryear{Wang \bgroup \em et al.\egroup }{2021}]{Advsim}
Jingkang Wang, Ava Pun, James Tu, Sivabalan Manivasagam, Abbas Sadat, Sergio Casas, Mengye Ren, and Raquel Urtasun.
\newblock Advsim: Generating safety-critical scenarios for self-driving vehicles.
\newblock In {\em Proceedings of the IEEE/CVF Conference on Computer Vision and Pattern Recognition}, pages 9909--9918, 2021.

\bibitem[\protect\citeauthoryear{Xu \bgroup \em et al.\egroup }{2022}]{Xu2022TrustworthyRL}
Mengdi Xu, Zuxin Liu, Peide Huang, Wenhao Ding, Zhepeng Cen, Bo~Li, and Ding Zhao.
\newblock Trustworthy reinforcement learning against intrinsic vulnerabilities: Robustness, safety, and generalizability.
\newblock {\em ArXiv}, abs/2209.08025, 2022.

\bibitem[\protect\citeauthoryear{Zhao \bgroup \em et al.\egroup }{2016}]{zhao2016accelerated}
Ding Zhao, Henry Lam, Huei Peng, Shan Bao, David~J LeBlanc, Kazutoshi Nobukawa, and Christopher~S Pan.
\newblock Accelerated evaluation of automated vehicles safety in lane-change scenarios based on importance sampling techniques.
\newblock {\em IEEE transactions on intelligent transportation systems}, 18(3):595--607, 2016.

\bibitem[\protect\citeauthoryear{Zhao \bgroup \em et al.\egroup }{2017}]{zhao2017accelerated}
Ding Zhao, Xianan Huang, Huei Peng, Henry Lam, and David~J LeBlanc.
\newblock Accelerated evaluation of automated vehicles in car-following maneuvers.
\newblock {\em IEEE Transactions on Intelligent Transportation Systems}, 19(3):733--744, 2017.

\bibitem[\protect\citeauthoryear{Zheng \bgroup \em et al.\egroup }{2022}]{zheng2022stackelberg}
Liyuan Zheng, Tanner Fiez, Zane Alumbaugh, Benjamin Chasnov, and Lillian~J Ratliff.
\newblock Stackelberg actor-critic: Game-theoretic reinforcement learning algorithms.
\newblock In {\em Proceedings of the AAAI conference on artificial intelligence}, volume~36, pages 9217--9224, 2022.

\bibitem[\protect\citeauthoryear{Zhou \bgroup \em et al.\egroup }{2019}]{zhou2019survey}
Yan Zhou, Murat Kantarcioglu, and Bowei Xi.
\newblock A survey of game theoretic approach for adversarial machine learning.
\newblock {\em Wiley Interdisciplinary Reviews: Data Mining and Knowledge Discovery}, 9(3):e1259, 2019.

\end{thebibliography}

\end{document}